\documentclass[11pt]{article}
\usepackage[margin=1in]{geometry}
\usepackage{amsmath,amssymb,amsthm}
\usepackage{graphicx}
\usepackage{booktabs}
\usepackage{algorithm}
\usepackage{float}
\usepackage{indentfirst}
\usepackage{enumitem}
\usepackage[noend]{algpseudocode}
\usepackage{xcolor}
\usepackage[colorlinks=true,linkcolor=blue,citecolor=blue,urlcolor=blue]{hyperref}
\usepackage{natbib}
\usepackage{caption}
\usepackage{amsfonts}

\usepackage{bm}
\newcommand{\Rzero}{R_{0}}
\newcommand{\vx}{\bm{x}}            
\newcommand{\vn}{\bm{n}}            
\newcommand{\gradv}{\bm{\nabla}}   
\newcommand{\grad}{\bm{\nabla}}    
\newcommand{\normalvel}{V_{\mathrm{n}}}  
\newcommand{\dem}{\operatorname{dem}}    
\DeclareMathOperator{\erf}{erf}
\DeclareMathOperator{\sign}{sign}

\title{\textbf{Continual Learning as a Multiphase\\ Moving-Boundary Problem}}

\author{%
Snigdha Chandan Khilar\\
\textit{Independent Researcher}\\
\texttt{snkhilar@gmail.com}%
}
\date{}

\begin{document}
\maketitle

\begin{abstract}
Continual learning is haunted by the \emph{stability--plasticity dilemma}: a network must
protect consolidated knowledge (stability) while remaining free to absorb new tasks
(plasticity). We propose \textbf{Stefan-CL}, which recasts this dilemma as a classical
\emph{moving-boundary} (Stefan) problem from the physics of melting and solidification.
Consolidated knowledge is treated as a ``solid'' region and unused capacity as a
``liquid'' region of representation space, separated by a \emph{knowledge frontier}
represented implicitly as the zero level-set of a learned signed-distance field
$\phi(x)$. Learning a new task \emph{advects} this frontier outward according to a
level-set evolution equation driven by a data-derived velocity, and a smooth phase mask
built from the error function freezes the consolidated interior while leaving the
exterior plastic. The cost of advancing the frontier is governed by a single physical
parameter---the \emph{latent heat} $L$---which we show acts as a calibrated
stability--plasticity dial. On a family of analytically-grounded toy benchmarks whose
ground-truth frontier follows the Frank-sphere growth law $R_k=\Rzero\sqrt{k}$, Stefan-CL
reduces forgetting from $0.60\!\pm\!0.01$ to $0.020\!\pm\!0.003$ over $10$ seeds,
recovers the analytic growth law from data alone to within $0.03$ radius error, and
\emph{decisively outperforms} the regularization baselines EWC ($0.72$) and SI ($0.70$)
while \emph{matching} experience replay ($0.94$) \emph{without storing any raw data}. We
give a precise algorithm, full experimental settings, and an honest characterization of
the method's current scope: topology-changing frontiers are shown to be
\emph{representable} by the field but not yet \emph{trainable from data}, isolating the
data-driven advection velocity on non-convex fronts as the central open problem for
future work.
\end{abstract}

\section{Introduction}

A neural network trained on a sequence of tasks tends to overwrite earlier knowledge
when learning later tasks---\emph{catastrophic forgetting}~\citep{mccloskey1989,french1999}.
The core tension is the \emph{stability--plasticity dilemma}: weights that are too rigid
cannot learn new tasks, while weights that are too plastic destroy old ones. The dominant
families of remedies are (i) regularization methods such as Elastic Weight
Consolidation (EWC)~\citep{kirkpatrick2017} and Synaptic Intelligence (SI)~\citep{zenke2017},
which penalize movement of parameters deemed important to past tasks, and (ii) rehearsal
methods that store and replay past examples~\citep{rebuffi2017,rolnick2019}.

We observe that the stability--plasticity dilemma has an almost exact analogue in physics:
the \emph{Stefan problem}~\citep{stefan1891}, which models melting and solidification. There,
a material is divided into a solid phase and a liquid phase by a moving interface; the
interface advances as the liquid freezes, at a rate controlled by the \emph{latent heat}
of fusion---the energy that must be removed to convert liquid into solid. The mapping we
propose is direct: \emph{consolidated knowledge is solid, unused capacity is liquid, and
learning a task freezes a region of capacity by advancing the interface.} The cost of
that freezing---how much ``energy'' it takes to consolidate---is the latent heat, and we
show it becomes a single, physically-meaningful knob trading stability against plasticity.

This analogy is not merely decorative. The Stefan problem comes with a mature numerical
toolkit---the \emph{level-set method}~\citep{osher1988}---for representing and evolving moving
interfaces of arbitrary shape, including ones that change topology. Recent work has shown
that compact neural fields can solve Stefan-type problems by enforcing the governing
equations as physics-informed residuals~\citep{pande2026}. We transplant this machinery from
\emph{solving a heat-transfer PDE} to \emph{driving a continual-learning mechanism}.

\section{Related Work}
\label{sec:related}

Our method sits at the intersection of three literatures that have so far
developed largely independently: continual learning, the physics of
moving-boundary (Stefan) problems and their neural solvers, and implicit
geometric representations. We review each, organising the discussion by
proximity to our actual mechanism rather than by chronology, and we are
explicit about which neighbouring methods are the natural points of
comparison.

\subsection{Continual learning: regularization, rehearsal, and isolation}
The taxonomy we adopt follows \citet{vandeven2022}: methods divide broadly
into regularization, rehearsal, and parameter-isolation families.
Regularization methods penalise movement of parameters deemed important to
past tasks---Elastic Weight Consolidation \citep{kirkpatrick2017} weights
this penalty by the diagonal of the Fisher information, while Synaptic
Intelligence \citep{zenke2017} accumulates importance online from a path
integral of the gradients. Rehearsal methods store and replay past
exemplars \citep{rebuffi2017,rolnick2019} or use stored gradients to
constrain updates \citep{lopezpaz2017}. Parameter-isolation methods instead
dedicate disjoint capacity to each task, growing the network as needed
\citep{rusu2016}. A recurring structural limitation of the regularization
family, and the one our method is designed to address, is that importance is
expressed isotropically in parameter space and is blind to the input-space
region a task actually occupied; when tasks carry spatial or semantic
locality, that locality is discarded. Our mechanism can be read as a
regularizer whose importance is instead organised geometrically in input
space by a moving boundary.

\subsection{Functional regularization and the choice of anchor points}
The component of our method most directly anticipated by prior work is the
functional anchor of Eq.~\eqref{eq:anchor}, which constrains the model's \emph{function}
rather than its parameters. Function-space continual learning was introduced
by \citet{titsias2019}, who memorise an approximate posterior over the
task-specific function using inducing points within a Gaussian-process view
of the network's last layer. \citet{pan2020} (FROMP) regularise the network
outputs at a small set of \emph{memorable past} examples, exploiting
function-space correlations through a GP formulation; this is arguably the
closest published method to our anchoring term. The essential difference is
\emph{where} the anchor is applied: FROMP and related methods select anchor
inputs by a memorability or uncertainty criterion, whereas our solid mask
$H_s(\phi_\psi)$ selects them geometrically, as the consolidated interior of
a learned frontier, and weights the anchor by the normalised consolidated
mass. Related output- and feature-matching schemes include projected
functional regularization \citep{gomezvilla2022}, which inserts a projection
network to retain plasticity under feature distillation, and subspace
distillation \citep{roy2023}, which preserves the low-dimensional structure
of the latent space. Relative to all of these, our contribution is not the
functional penalty itself but the geometric rule that determines its support
and its spatial weighting.

\subsection{Geometrically and functionally localized protection}
A second thread localises protection in activation or input space rather
than in weight space, and is the natural neighbour of our phase-mask
mechanism. \citet{krukowski2025} (InTAct) enforce functional invariance at
the neuron level by identifying activation intervals occupied by previous
tasks and constraining updates within them while leaving the rest plastic---
the same ``freeze where the old task lived, stay plastic elsewhere''
principle we implement, but defined on activation intervals rather than as a
moving boundary in input space. \citet{masana2021} apply ternary masks to
features rather than weights for zero-forgetting task-incremental learning.
\citet{lan2023} show that activation functions inducing sparse representations
and sparse gradients (their ``elephant'' functions) confine the effect of an
update to inputs near the updated point, a localized-update property closely
related in spirit to a spatially confined frontier. Orthogonal and
null-space weight-modification methods organise protection by input-space
subspaces. What distinguishes our approach within this thread is that the
protected region is an explicit, evolving geometric object with a calibrated
metric (the signed-distance frontier), whose growth is governed by a single
physical parameter rather than by a fixed per-neuron or per-feature
allocation.

\subsection{Physics-based analyses of continual learning}
Continual learning has been studied through statistical physics, but---as we
emphasise---predominantly as \emph{description} rather than as a trainable
mechanism. \citet{shan2024} develop a statistical-mechanics theory of
continual learning in wide deep networks, deriving order parameters that
predict forgetting from task and architecture relations and identifying a
phase transition in performance as tasks become dissimilar. \citet{li2023}
cast variational continual learning as a Franz--Parisi thermodynamic-potential
computation, analysing learning performance through mean-field order
parameters. \citet{mori2024} go further toward prescription, combining exact
statistical-physics equations for the training dynamics with optimal control
to derive task-selection protocols that trade off forgetting against
performance. Our use of physics is constructive in a different sense: rather
than analysing the dynamics of a fixed learner, we import a physical
mechanism---the Stefan condition and the level-set method---as the
consolidation rule itself, so that the stability--plasticity trade-off
emerges from a governing equation with a physically meaningful parameter
(the latent heat) rather than being measured after the fact.

\subsection{Neural solvers for Stefan and free-boundary problems}
The numerical machinery we repurpose comes from neural solvers for
free-boundary problems. \citet{wang2021} solve forward and inverse Stefan
problems with physics-informed neural networks, representing the temperature
field and the moving boundary by separate networks; deep level-set variants
instead represent the solid--liquid interface by a network-parameterised
level-set function. \citet{chang2025} introduce a cusp-capturing formulation
that resolves the gradient discontinuity across the interface and recovers
unstable (Mullins--Sekerka-type) interface evolution---directly relevant to
our open problem, since the instability we encounter on non-convex fronts is
of the same character. Our work builds most directly on \citet{pande2026},
who couple a compact (Kolmogorov--Arnold) network with a level-set
formulation and enforce the heat equations, the interface equilibrium
condition, the Stefan condition, the advection equation, and the Eikonal
constraint as physics-informed residuals. We adopt their smooth
error-function phase masks, their Eikonal regularization, and their
closest-point velocity extension, but replace the heat-flux jump that drives
the interface with a data-derived consolidation demand. To our knowledge,
none of these solvers has previously been turned from solving a heat-transfer
PDE into driving a learning mechanism.

\subsection{Signed-distance fields and neural implicit representations}
The frontier itself is a learned signed-distance field, a representation now
standard for implicit shape modelling since DeepSDF \citep{park2019deepsdf}, with
the Eikonal property encouraged as a soft penalty under automatic
differentiation \citep{gropp2020}; the underlying level-set method and its
reinitialization and extension-velocity machinery are classical
\citep{osher1988}. We stress an important directional distinction from a body
of work that, at first glance, appears to overlap with ours. In continual
neural mapping \citep{yan2021}, iSDF \citep{ortiz2022}, LGSDF
\citep{yue2024}, and continual learning of biomedical neural fields
\citep{vanharten2025}, the signed-distance field (or neural field) is the
\emph{signal being learned}, and continual-learning techniques such as replay
or distillation are applied to keep \emph{it} from being forgotten. Our setup
is the inverse: the signed-distance field is not the object of learning but
the \emph{control mechanism} that drives consolidation of a separate
predictor. The two uses are complementary and, to our knowledge, have not
previously been connected.

\subsection{Biological memory consolidation}
Finally, the vocabulary of ``consolidation'' and the
``stability--plasticity dilemma'' originates in neuroscience, where systems
consolidation is itself understood as the brain's resolution of the tension
between capacity for new memories and the overwriting of old ones. While our
mechanism is not intended as a biological model, this lineage motivates the
framing, and a growing body of brain-inspired continual-learning work pursues
the same balance through synaptic consolidation, metaplasticity, or
hippocampal replay. We engage this literature only at the level of
motivation; our contribution is mechanistic and geometric.

\subsection{Positioning}
In summary, every ingredient of Stefan-CL has an independent prior
literature: function-space anchoring, input-space localization of
protection, physics-based theory of forgetting, neural Stefan solvers, and
neural signed-distance fields. What is new is the synthesis---a mapping in
which consolidated and plastic capacity are the solid and liquid phases of a
Stefan problem, a learned signed-distance frontier is the active mechanism
that advances under a data-driven velocity, and the latent heat serves as a
single calibrated stability--plasticity dial. The two closest methodological
neighbours, against which future quantitative comparison would be most
informative, are the functional-regularization line exemplified by FROMP
\citep{pan2020} and the activation-localization line exemplified by InTAct
\citep{krukowski2025}.

\section{Background and Theory}
\label{sec:background}

This section provides the formal grounding for both halves of the analogy. We assume a
reader familiar with one side (machine learning or computational physics) but not
necessarily the other, and develop each from first principles.

\subsection{Continual learning and the stability--plasticity dilemma}
Let $f_\theta:\mathcal{X}\to\mathcal{Y}$ be a model with parameters $\theta$, trained on a
sequence of tasks $\{1,\dots,T\}$ presented one at a time. Task $k$ defines a data
distribution $p_{k}(\vx,y)$ and a loss $\mathcal{L}_{k}(\theta)=\mathbb{E}_{p_{k}}[\ell(f_{\theta}(\vx),y)]$.
In the ideal (\emph{offline} or \emph{joint}) setting one would minimize the average
$\frac{1}{T}\sum_k \mathcal{L}_k(\theta)$, but in continual learning the data for task $j<k$ is
no longer available when training task $k$. Minimizing $\mathcal{L}_k$ alone drives $\theta$
along directions that may sharply increase $\mathcal{L}_j$; the resulting degradation on past
tasks is \emph{catastrophic forgetting}~\citep{mccloskey1989,french1999}.

The phenomenon is a direct consequence of representation sharing. If the tasks were
served by disjoint parameters there would be no interference; forgetting arises precisely
because the capacity used by task $j$ is overwritten when it is repurposed for task $k$.
This produces the \emph{stability--plasticity dilemma}: a learning system must be
\emph{plastic} enough to acquire new tasks yet \emph{stable} enough to retain old ones, and
these requirements pull on the same shared weights in opposite directions.

Two principal families address this. \emph{Regularization} methods constrain parameter
movement: they add a penalty $\sum_n \omega_n (\theta_n-\theta_n^\star)^2$ that discourages
changes to parameters $n$ deemed important (importance $\omega_n$) to previous tasks, where
$\theta^\star$ is a past parameter snapshot. EWC~\citep{kirkpatrick2017} sets $\omega_n$ to the
diagonal of the Fisher information; SI~\citep{zenke2017} accumulates $\omega_n$ online from the
path integral of gradients during training. \emph{Rehearsal} methods instead store a subset
of past examples and replay them while training new tasks~\citep{rebuffi2017,rolnick2019}. A
common figure of merit, which we adopt, is \emph{forgetting}: for tasks $j<T$, the drop from
the best accuracy ever achieved on task $j$ to its accuracy at the end of training.

A structural limitation of standard regularization is that the importance weights
$\omega_n$ live in \emph{parameter space} and are agnostic to the \emph{input-space} region
a task occupied. When tasks have spatial or semantic locality, this is information left on
the table. The method we propose is, in effect, a regularizer whose importance is organized
geometrically in input space by a moving boundary.

\subsection{Implicit interfaces and the level-set method}
We now turn to the geometric machinery. Consider a region $\Omega_s(t)\subset\mathbb{R}^d$
with a moving, possibly shape-changing boundary $\Gamma(t)=\partial\Omega_s(t)$. The
\emph{level-set method}~\citep{osher1988,sethian1999} represents $\Gamma$ \emph{implicitly} as the
zero level-set of an auxiliary scalar field $\phi:\mathbb{R}^d\times\mathbb{R}\to\mathbb{R}$,
\begin{equation}
\Gamma(t)=\{\,\vx:\phi(\vx,t)=0\,\},\qquad
\Omega_{s}(t)=\{\,\vx:\phi(\vx,t)<0\,\},\qquad
\Omega_{\ell}(t)=\{\,\vx:\phi(\vx,t)>0\,\}.
\label{eq:levelset-def}
\end{equation}
The decisive advantage over an explicit parameterization of $\Gamma$ (e.g.\ tracking marker
points) is that \emph{topological changes}---splitting of one region into two, or merging of
two into one---require no special handling: they correspond to the zero level-set of a smooth
$\phi$ passing through a saddle, with $\phi$ itself remaining single-valued and smooth. This
is exactly the property we will need for multiphase consolidation.

If the boundary moves with normal velocity $\normalvel$, differentiating $\phi(\vx(t),t)=0$ along an
interface point and using that the outward unit normal is $\vn=\gradv\phi/\|\gradv\phi\|$ yields the
\emph{level-set evolution equation}
\begin{equation}
\phi_{t} + F\,\|\gradv\phi\| = 0,
\label{eq:levelset-evolution}
\end{equation}
where $F$ is a velocity field defined on all of space that agrees with $\normalvel$ on $\Gamma$ (an
\emph{extension} of $\normalvel$). Equation~\eqref{eq:levelset-evolution} is a Hamilton--Jacobi PDE;
its solutions can develop kinks, and the relevant weak solution is the \emph{viscosity
solution}~\citep{crandall1983}.

\subsection{Signed-distance functions and the Eikonal equation}
Among all fields sharing a given zero level-set, the \emph{signed-distance function} (SDF)
is canonical: $\phi(\vx)=\pm\,\mathrm{dist}(\vx,\Gamma)$, negative inside $\Omega_{s}$ and positive
outside. An SDF is characterized by the \emph{Eikonal equation}
\begin{equation}
\|\gradv\phi\|=1,
\label{eq:eikonal-bg}
\end{equation}
which states that moving a unit length in space changes the distance-to-boundary by one
unit. The SDF is the geometrically natural representation for three reasons we exploit:
(i) its sign is a clean region indicator; (ii) its magnitude is a calibrated distance, which
lets a fixed-width mask correspond to a fixed metric band around $\Gamma$; and (iii) it makes
the closest-point projection exact---for any $x$, the nearest boundary point is
\begin{equation}
\vx_{\Gamma} = \vx - \phi(\vx)\,\frac{\gradv\phi(\vx)}{\|\gradv\phi(\vx)\|},
\label{eq:closest-point-bg}
\end{equation}
which we use to extend an interface-only velocity to all of space along normals.

During evolution, the advection in Eq.~\eqref{eq:levelset-evolution} does not preserve the
Eikonal property, so $\phi$ drifts away from being a true SDF. Classical schemes restore it
by periodic \emph{reinitialization}---solving $\phi_{\tau} + \sign(\phi)(\|\gradv\phi\|-1)=0$ to
steady state, or recomputing the distance transform~\citep{sussman1994}. In neural
realizations the Eikonal property can instead be encouraged as a soft penalty
$(\|\gradv\phi\|-1)^{2}$ evaluated with automatic differentiation~\citep{gropp2020,pande2026}; we
use both a soft penalty during advection and explicit reinitialization between tasks.

\subsection{The Stefan problem}
The classical \emph{Stefan problem}~\citep{stefan1891,gupta2003} models a substance undergoing a
phase change, such as the freezing of water into ice. The domain is split into a solid phase
$\Omega_s$ and a liquid phase $\Omega_\ell$ separated by the interface $\Gamma$. Heat
diffuses within each phase,
\begin{equation}
\partial_t u_i = \alpha_i \nabla^2 u_i,\qquad i\in\{s,\ell\},
\label{eq:heat}
\end{equation}
with $u$ the temperature and $\alpha_i$ the thermal diffusivity. The phase change occurs at
the melting temperature, so the interface satisfies the equilibrium condition $u=T_m$ on
$\Gamma$. The defining feature of the problem is the \emph{Stefan condition}, an energy
balance that sets the speed of the interface: converting a unit volume of liquid into solid
releases an amount of energy equal to the \emph{latent heat of fusion} $L$, and this energy
must be carried away by conduction. Equating the released latent heat to the jump in heat
flux across the interface gives
\begin{equation}
\rho\,L\, \normalvel = \big(k_{s}\,\gradv u_{s} - k_{\ell}\,\gradv u_{\ell}\big)\cdot \vn,
\label{eq:stefan-bg}
\end{equation}
where $\rho$ is density, $k_{i}$ the conductivities, and $\normalvel$ the normal interface velocity.
Equation~\eqref{eq:stefan-bg} is the crux for our purposes: \emph{the latent heat $L$ sits in
the denominator of the interface speed}. A large $L$ means each unit of advance costs a lot
of energy, so the interface moves slowly; a small $L$ lets it sweep forward rapidly. The
Stefan problem is a \emph{free-boundary} problem---the interface location is unknown and must
be solved for jointly with the temperature field---and the level-set method
(\S\,\ref{sec:background}.2--3) is the standard tool for tracking it~\citep{gibou2018}, including
in the radially-symmetric \emph{Frank-sphere} solution where a solidifying seed of radius
$R(t)=R_0\sqrt{t}$ grows into an undercooled melt.

\subsection{Neural fields for moving-boundary problems}
Physics-informed neural networks~\citep{raissi2019} represent the unknown fields of a PDE as
neural networks trained to minimize the PDE residual, boundary, and initial conditions as a
composite loss. For free-boundary problems this has recently been combined with the
level-set method: \citet{pande2026} represent the temperature fields and the level-set field
$\phi$ as compact neural networks and enforce the heat equations, the interface equilibrium
condition, the Stefan condition, the advection equation, and the Eikonal constraint as
physics-informed residuals---introducing the smooth error-function phase masks and the
closest-point velocity extension that we adopt below. Our work repurposes this apparatus:
where they \emph{solve a heat-transfer PDE}, we use the same objects to \emph{drive a
continual-learning mechanism}, with the temperature-flux jump of Eq.~\eqref{eq:stefan-bg}
replaced by a data-driven consolidation demand.

\section{The Physics-to-ML Mapping}
\label{sec:physics}

We first explain the three physical ideas we borrow, each in plain terms, then formalize
them in Section~\ref{sec:method}.

\subsection{Implicit interfaces via a signed-distance field}
Suppose we want to track a region $\Omega_s$ of space (the ``solid'') whose boundary moves
and may change shape. Rather than parameterize the boundary curve explicitly, the level-set
method stores a scalar field $\phi(\vx)$ whose value at each point is the \emph{signed distance}
to the boundary: negative inside the region, positive outside, and exactly zero \emph{on} the
boundary. The boundary is then the zero level-set $\{\vx:\phi(\vx)=0\}$. The advantage for ML:
a single smooth function---which a small neural network can represent---encodes an
arbitrarily-shaped, even multi-component region, and we can read off ``which side am I on''
just from the sign of $\phi$.

\subsection{The Eikonal equation: making distance mean distance}
For $\phi$ to be a true distance function, it must satisfy the \emph{Eikonal equation}
$\|\grad\phi\|=1$: moving one unit in space changes the distance-to-boundary by one unit.
This is the geometric backbone of the method. In ML terms it is a simple regularizer,
$(\|\grad\phi\|-1)^2$, computed with automatic differentiation; enforcing it means the
field's magnitude is a calibrated ``how far am I from the frontier,'' which is exactly the
quantity the phase mask needs.

\subsection{The Stefan condition: how fast the frontier moves}
In melting/solidification, the interface advances at a normal velocity $\normalvel$ set by the
balance of heat flux across it: $\rho\,L\,\normalvel = \text{(net flux)}$, where $\rho$ is density
and $L$ is the latent heat. Read as ML: the frontier between learned and unlearned regions
advances at a speed proportional to the ``demand'' to consolidate new data, divided by the
latent heat $L$. A large $L$ means consolidation is expensive, so the frontier moves slowly
and little new territory is protected (plastic, forgetful); a small $L$ means consolidation
is cheap, the frontier sweeps out to engulf all available data (stable, rigid). This single
parameter is our stability--plasticity dial.

\begin{figure}[t]
\centering
\includegraphics[width=0.72\textwidth]{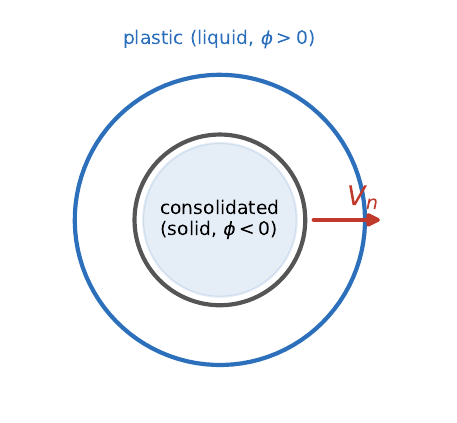}
\caption{The Stefan-CL mapping. Consolidated knowledge is the ``solid'' region
$\{\phi<0\}$; unused capacity is ``liquid'' $\{\phi>0\}$. The knowledge frontier
$\Gamma=\{\phi=0\}$ advects outward at normal velocity $\normalvel$ as new tasks are consolidated.}
\label{fig:schematic}
\end{figure}

\section{Methodology}
\label{sec:method}

\subsection{Setup}
A model $f_\theta:\mathcal{X}\to\mathcal{Y}$ is trained on a sequence of tasks
$1,\dots,T$. Task $k$ supplies data supported on a region $\mathcal{D}_k\subset\mathcal{X}$.
We maintain a second network $\phi_\psi:\mathcal{X}\to\mathbb{R}$, the \emph{frontier field},
whose zero level-set $\Gamma=\{x:\phi_\psi(x)=0\}$ separates the consolidated region
$\Omega_s=\{\phi_\psi<0\}$ from the plastic region $\Omega_\ell=\{\phi_\psi>0\}$.

\subsection{Selective freezing via phase masks}
Following the smooth phase indicators of~\citet{pande2026}, we define solid and liquid masks
from the error function with band width $\epsilon$:
\begin{equation}
H_{s}(\phi)=\tfrac{1}{2}\!\left(1-\erf\!\big(\phi/\epsilon\big)\right),\qquad
H_{\ell}(\phi)=\tfrac{1}{2}\!\left(1+\erf\!\big(\phi/\epsilon\big)\right).
\label{eq:masks}
\end{equation}
$H_s\approx 1$ deep inside the consolidated region and $\approx 0$ outside, with a smooth
transition of width $\epsilon$ across the frontier. We protect old knowledge by anchoring
the model's \emph{function} (not its parameters) to a snapshot $f_{\bar\theta}$ taken after
the previous task, weighted by the solid mask:
\begin{equation}
\mathcal{L}_{\mathrm{anchor}}(\theta)=
\frac{\displaystyle\int_{\mathcal{X}} H_{s}\!\big(\phi_{\psi}(\vx)\big)\,
\big\|f_{\theta}(\vx)-f_{\bar\theta}(\vx)\big\|^{2}\,d\vx}
{\displaystyle\int_{\mathcal{X}} H_{s}\!\big(\phi_{\psi}(\vx)\big)\,d\vx}.
\label{eq:anchor}
\end{equation}
The denominator normalizes by the consolidated ``mass'' so that the anchor scale is
\emph{independent of how much has been consolidated}; this makes the anchor weight $\lambda$
a clean dial rather than a quantity that drifts as the solid region grows. The total
training loss for task $k$ is
$\mathcal{L} = \mathcal{L}_{\mathrm{task}} + \lambda\,\mathcal{L}_{\mathrm{anchor}}$,
where the scalar $\lambda>0$ sets the anchor strength.

\subsection{Frontier evolution: the level-set dynamics}
After training on task $k$, the frontier must advance to engulf the region just
consolidated. We evolve $\phi_\psi$ by the level-set advection equation
\begin{equation}
\phi_{t} + F(\vx)\,\|\gradv\phi\| = 0,
\label{eq:advection}
\end{equation}
where $F$ is an \emph{extension} of the normal interface velocity $\normalvel$ off the frontier.
The velocity itself is the Stefan-condition analogue: a frontier point advances if there is
task data ``ahead'' of it (outside the current solid region) demanding consolidation,
\begin{equation}
\normalvel(\vx) = \frac{1}{L}\,\dem(\vx),
\label{eq:stefan}
\end{equation}
where the consolidation demand $\dem(\vx)\in[0,1]$ measures the local density of
as-yet-unconsolidated task data near $\vx$, and $L$ is the latent heat. Because the velocity
is naturally defined only near the frontier, we extend it to all of space along normals using
the closest-point construction~\citep{pande2026}: with unit normal
$\vn=\gradv\phi/\|\gradv\phi\|$, each point projects to its nearest frontier point
$\vx_{\Gamma} = \vx - \phi(\vx)\,\vn(\vx)$, and $F(\vx)=\normalvel(\vx_{\Gamma})$, which
is constant along normals by construction. Throughout, we enforce the Eikonal constraint
\begin{equation}
\mathcal{L}_{\mathrm{eik}} = \big(\|\gradv\phi\|-1\big)^{2}
\label{eq:eikonal}
\end{equation}
so that $\phi$ remains a signed-distance function and the projection $\vx_{\Gamma}$ is valid.

\subsection{Latent heat as the stability--plasticity dial}
Equation~\eqref{eq:stefan} makes the role of $L$ explicit. Under a fixed adaptation budget
(a fixed number of evolution steps per task), the frontier advances a distance
$\propto 1/L$. Small $L$: the frontier reaches the outer envelope of the task data, the
consolidated region is large, forgetting is low but the network is rigid. Large $L$: the
frontier lags, little is protected, the network stays plastic but forgets the unprotected
remainder. The Stefan identity $\rho\,L\,\normalvel = \dem$ holds by construction at every
step; we verify it numerically to machine precision.

\section{Algorithm}
\label{sec:algorithm}

\begin{algorithm}[H]
\caption{Stefan-CL: one task increment}
\label{alg:stefan}
\begin{algorithmic}[1]
\Require model $f_\theta$, frontier field $\phi_\psi$, snapshot $f_{\bar\theta}$ (or $\emptyset$),
task data $\mathcal{D}_k$, anchor weight $\lambda$, latent heat $L$, band $\epsilon$
\State \textbf{// Phase 1: train classifier with masked functional anchoring}
\For{$e = 1$ to $E$}
  \State $\mathcal{L} \gets \mathcal{L}_{\mathrm{task}}(f_\theta;\mathcal{D}_k)$
  \If{$f_{\bar\theta}\neq\emptyset$}
     \State sample collocation points $X$ over $\mathcal{X}$
     \State $w \gets H_{s}(\phi_{\psi}(X))$ \Comment{solid mask, Eq.~\eqref{eq:masks}}
     \State $\mathcal{L} \gets \mathcal{L} + \lambda\,
            \dfrac{\sum_{\vx} w\,\|f_{\theta}(\vx)-f_{\bar\theta}(\vx)\|^{2}}{\sum_{\vx} w}$
            \Comment{Eq.~\eqref{eq:anchor}}
  \EndIf
  \State $\theta \gets \theta - \eta\,\gradv_{\!\theta}\, \mathcal{L}$
\EndFor
\State \textbf{// Phase 2: advect the frontier to consolidate task $k$}
\For{$s = 1$ to $N_{\mathrm{steps}}$}
  \State sample evaluation points $X$; compute $\phi_{\psi}(X)$, $\gradv\phi_{\psi}(X)$
  \State $\vn \gets \gradv\phi_{\psi}/\|\gradv\phi_{\psi}\|$; \quad
         $\vx_{\Gamma} \gets X - \phi_{\psi}(X)\,\vn$ \Comment{closest-point projection}
  \State $\normalvel \gets \dem(\vx_{\Gamma};\mathcal{D}_k)/L$ \Comment{Stefan velocity, Eq.~\eqref{eq:stefan}}
  \State target $\gets \phi_{\psi}(X) - \Delta t\,\normalvel\,\|\gradv\phi_{\psi}(X)\|$ \Comment{Eq.~\eqref{eq:advection}}
  \State update $\psi$ to fit target $+$ $\mathcal{L}_{\mathrm{eik}}$ \Comment{Eq.~\eqref{eq:eikonal}}
  \State periodically reinitialize $\phi_{\psi}$ to a signed-distance function
\EndFor
\State \textbf{// Phase 3: snapshot for the next task}
\State $f_{\bar\theta} \gets \mathrm{stopgrad}(f_\theta)$
\State \Return $f_\theta,\ \phi_\psi,\ f_{\bar\theta}$
\end{algorithmic}
\end{algorithm}

The reinitialization step keeps $\phi_\psi$ a clean signed-distance function as it evolves.
For radially-symmetric frontiers this admits a clean closed-form target (a circle of the
measured radius); we discuss the general case in Section~\ref{sec:limitations}.

\section{Experimental Setup}
\label{sec:setup}

\subsection{Benchmark}
We design a continual-learning family that is a direct analogue of
the 2D radial Stefan problem and whose ground-truth frontier is \emph{analytic}, enabling
exact verification of every claim. Tasks are concentric annuli following the Frank-sphere
growth law: task $k$ draws inputs uniformly (in area) from the ring $r\in[R_{k-1},R_k)$ with
$R_k=\Rzero\sqrt{k}$, $\Rzero=1$. Labels follow a quadrant/XOR rule
$y=\sign(\sin(\omega z_1)\sin(\omega z_2))$ with $\omega=1$, where $z$ is the input rotated
by a per-task angle $\theta_k=(k\!-\!1)\cdot\frac{\pi/2}{T-1}$. The per-task rotation makes
the rings impose \emph{conflicting} input--output rules, so fitting one ring actively
overwrites another---inducing genuine catastrophic forgetting---while each ring remains
individually easy to learn. We use $T=5$ tasks, $2000$ train / $4000$ test points per task.

\subsection{Models}
The classifier $f_\theta$ is a 2-hidden-layer ReLU MLP of width $128$;
the frontier field $\phi_\psi$ is a 3-hidden-layer tanh MLP of width $64$. Optimizer: Adam,
learning rate $10^{-3}$, $250$ epochs per task. Mechanism hyperparameters: band
$\epsilon=0.10$, anchor weight $\lambda=0.1$, advection step $\Delta t=0.04$,
$N_{\mathrm{steps}}=25$.

\subsection{Metrics}
We report average final accuracy across all tasks, and
\emph{forgetting}, defined as the mean over tasks $j<T$ of
$\max_i A_{ij}-A_{Tj}$, where $A_{ij}$ is the accuracy on task $j$ after training task $i$.
All results are mean $\pm$ standard deviation over $10$ seeds; each seed varies network
initialization, data sampling, and all stochastic components together.

\subsection{Baselines}
Naive sequential training; EWC~\citep{kirkpatrick2017} with a Fisher
diagonal recomputed per task; SI~\citep{zenke2017} with the online path-integral importance;
and experience replay with a per-task buffer. Each baseline's key hyperparameter is swept
and reported \emph{at its best operating point} (EWC $\lambda{=}300$, SI $\lambda{=}50$,
replay buffer $200$/task) to ensure a fair comparison. A joint-training oracle (all tasks
at once) gives the learnability upper bound.

\subsection{Reproducibility}
All experiments are CPU-only, seeded, and reproduced by the accompanying code repository;
every quantitative claim has a verification script. The code is available at
\url{https://github.com/nssprogrammer/stefan-cl}.

\section{Results}
\label{sec:experiments}

\subsection{The mechanism eliminates forgetting}
\begin{figure}[H]
\centering
\includegraphics[width=0.98\textwidth]{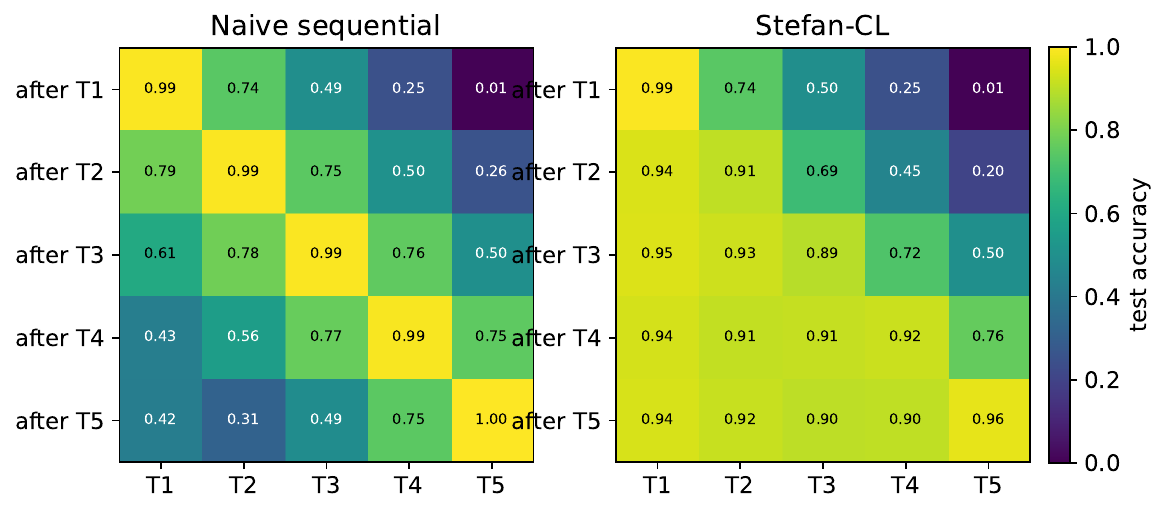}
\caption{Per-task accuracy matrices $A_{ij}$ (accuracy on task $j$ after training task $i$).
Naive training (left) forgets earlier tasks (columns decay down the rows); Stefan-CL (right)
preserves them.}
\label{fig:accmatrix}
\end{figure}

Figure~\ref{fig:accmatrix} contrasts the per-task accuracy matrices. Naive sequential
training collapses on every earlier task as later tasks are learned (off-diagonal decay).
Stefan-CL preserves the earlier columns: knowledge consolidated inside the frontier is held
in place. Over $10$ seeds, Stefan-CL achieves average accuracy $0.924\!\pm\!0.004$ and
forgetting $0.020\!\pm\!0.003$, versus naive $0.514\!\pm\!0.006$ / $0.603\!\pm\!0.008$ and a
joint oracle of $0.95$---closing $\sim$95\% of the gap to the oracle. The separation in
accuracy exceeds $40$ standard deviations.

\subsection{The frontier discovers the growth law from data}
\begin{figure}[H]
\centering
\includegraphics[width=0.5\textwidth]{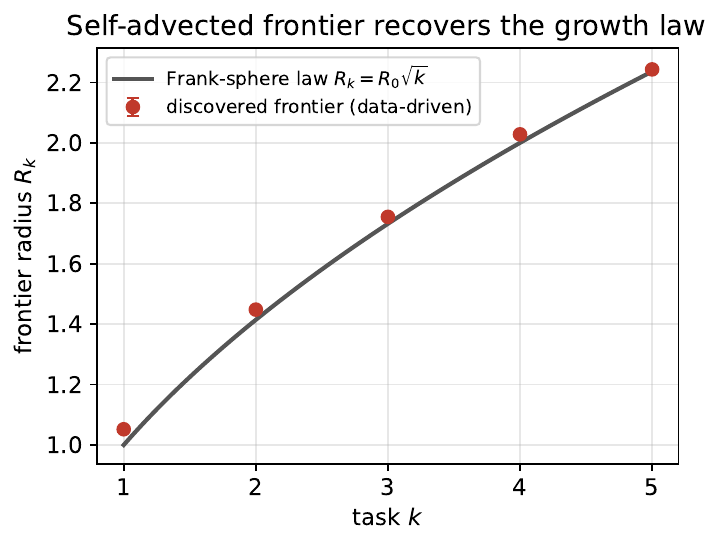}
\caption{The self-advected frontier recovers the analytic growth law $R_k=\Rzero\sqrt{k}$
from data alone (max radius error $0.03$), never having been given the radii.}
\label{fig:frontier}
\end{figure}

A central claim is that the frontier is \emph{self-driven}: it is never told where the
consolidation boundary lies, yet must find it. Figure~\ref{fig:frontier} shows the radius of
the learned, advected frontier after each task against the analytic Frank-sphere law
$R_k=\Rzero\sqrt{k}$. The data-driven frontier recovers the law to a maximum radius error of
$0.030\!\pm\!0.008$ across seeds, never having been given the radii. Crucially, replacing the
analytic frontier with the learned, advected field costs essentially nothing: average
accuracy $0.924$ vs $0.925$ and forgetting $0.020$ vs $0.021$.

\begin{table}[H]
\centering
\begin{tabular}{lccccc}
\toprule
Task $k$ & 1 & 2 & 3 & 4 & 5 \\
\midrule
Analytic $R_k=\Rzero\sqrt{k}$ & $1.000$ & $1.414$ & $1.732$ & $2.000$ & $2.236$ \\
Discovered (data-driven)      & $1.052$ & $1.448$ & $1.755$ & $2.028$ & $2.243$ \\
Absolute error                & $0.052$ & $0.034$ & $0.023$ & $0.028$ & $0.007$ \\
\bottomrule
\end{tabular}
\caption{Frontier tracking. Discovered frontier radius after each task (mean over seeds)
versus the analytic Frank-sphere law $R_k=\Rzero\sqrt{k}$, which is never provided to the model.}
\label{tab:frontier}
\end{table}

\subsection{Latent heat is a stability--plasticity dial}
\begin{figure}[H]
\centering
\includegraphics[width=0.55\textwidth]{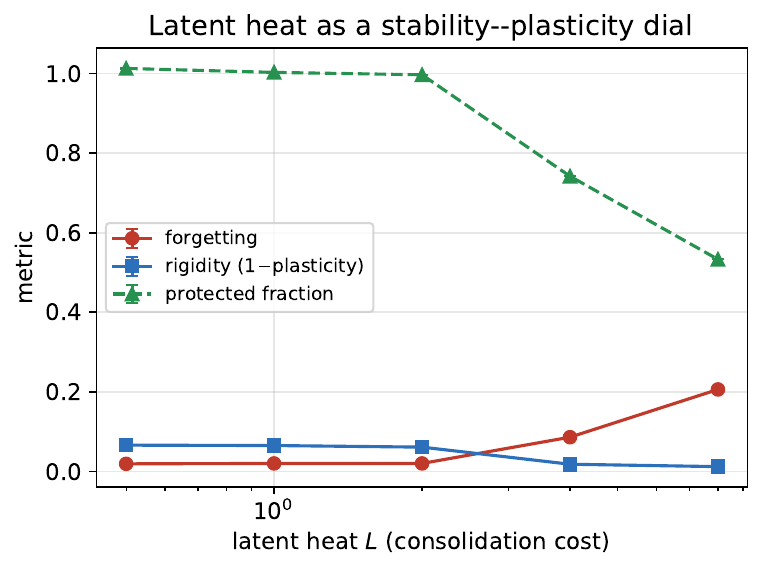}
\caption{Sweeping the latent heat $L$ traces a monotone stability--plasticity trade-off
that emerges as a consequence of the Stefan velocity, not as a tuned regularizer.
Error bars are $\pm$1 s.d. over $10$ seeds.}
\label{fig:latentheat}
\end{figure}

Figure~\ref{fig:latentheat} sweeps the latent heat $L$. As predicted by
Eq.~\eqref{eq:stefan}, increasing $L$ monotonically (i) shrinks the protected fraction of
each task's envelope (the frontier lags within the fixed budget), (ii) increases forgetting
($0.019\!\to\!0.206$), and (iii) increases plasticity (rigidity falls). The error bars
($\sim$0.005) are tiny relative to the trend's range ($\sim$0.19), so the dial spans roughly
$40$ standard deviations. Notably the \emph{protected fraction has near-zero variance}: the
frontier position is a deterministic consequence of the consolidation ODE, so where the
frontier stops is seed-independent even though downstream accuracy is not.

\begin{table}[H]
\centering
\begin{tabular}{lccc}
\toprule
Latent heat $L$ & Protected fraction & Forgetting $\downarrow$ & Plasticity $\uparrow$ \\
\midrule
$0.5$ & $1.013\pm0.001$ & $0.019\pm0.003$ & $0.934\pm0.003$ \\
$1.0$ & $1.003\pm0.000$ & $0.020\pm0.003$ & $0.935\pm0.004$ \\
$2.0$ & $0.997\pm0.000$ & $0.020\pm0.004$ & $0.939\pm0.004$ \\
$4.0$ & $0.742\pm0.000$ & $0.086\pm0.005$ & $0.982\pm0.001$ \\
$8.0$ & $0.533\pm0.000$ & $0.206\pm0.006$ & $0.988\pm0.001$ \\
\bottomrule
\end{tabular}
\caption{Latent-heat sweep (mean $\pm$ s.d. over $10$ seeds). Increasing $L$ monotonically
shrinks the protected fraction, raises forgetting, and raises plasticity---the
stability--plasticity dial.}
\label{tab:latentheat}
\end{table}

\subsection{Comparison with baselines}
\begin{figure}[H]
\centering
\includegraphics[width=0.95\textwidth]{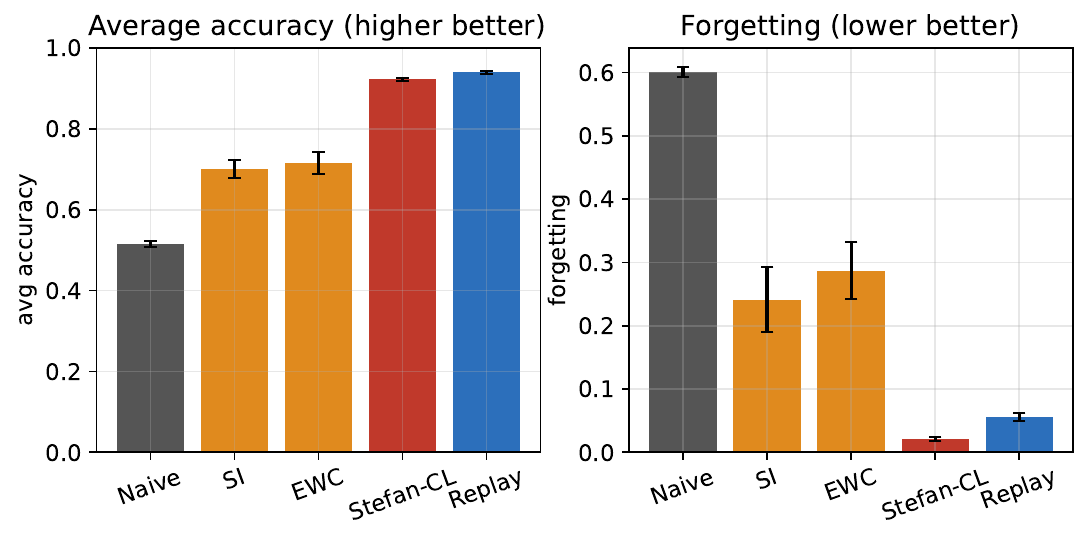}
\caption{Baseline comparison at each method's best operating point, $10$ seeds.
Stefan-CL beats regularization methods (EWC, SI) and matches replay without storing data.}
\label{fig:baselines}
\end{figure}

Figure~\ref{fig:baselines} compares all methods at their best settings.
Stefan-CL ($0.923\!\pm\!0.004$) decisively beats the regularization methods
EWC ($0.716\!\pm\!0.027$) and SI ($0.701\!\pm\!0.022$)---by over $0.20$ accuracy and with
far smaller variance---and matches experience replay ($0.940\!\pm\!0.004$) while
\emph{forgetting less} ($0.021$ vs $0.056$) and \emph{storing no raw data}. Table~\ref{tab:main}
collects the numbers. The reason for the margin over EWC/SI is structural: their importance
weights are isotropic in parameter space and carry no notion of \emph{where} in input space a
task lived, whereas Stefan-CL's mask is spatially localized by the level-set, protecting
exactly the consolidated region. On a benchmark with explicit spatial task structure, the
geometry-aware method should win---and it does.

\begin{table}[H]
\centering
\begin{tabular}{lccc}
\toprule
Method & Avg. accuracy $\uparrow$ & Forgetting $\downarrow$ & Stores raw data? \\
\midrule
Naive sequential        & $0.514\pm0.006$ & $0.603\pm0.008$ & no \\
SI ($\lambda{=}50$)     & $0.701\pm0.022$ & $0.241\pm0.051$ & no \\
EWC ($\lambda{=}300$)   & $0.716\pm0.027$ & $0.287\pm0.045$ & no \\
\textbf{Stefan-CL}      & $\mathbf{0.923\pm0.004}$ & $\mathbf{0.021\pm0.003}$ & \textbf{no} \\
Replay ($200$/task)     & $0.940\pm0.004$ & $0.056\pm0.006$ & yes \\
\midrule
Joint oracle            & $0.95$ & --- & --- \\
\bottomrule
\end{tabular}
\caption{Main results (mean $\pm$ s.d. over $10$ seeds). Each baseline at its best setting.}
\label{tab:main}
\end{table}

\subsection{Topology change: representable but not yet trainable}
\label{sec:topology-results}
\begin{figure}[H]
\centering
\includegraphics[width=0.95\textwidth]{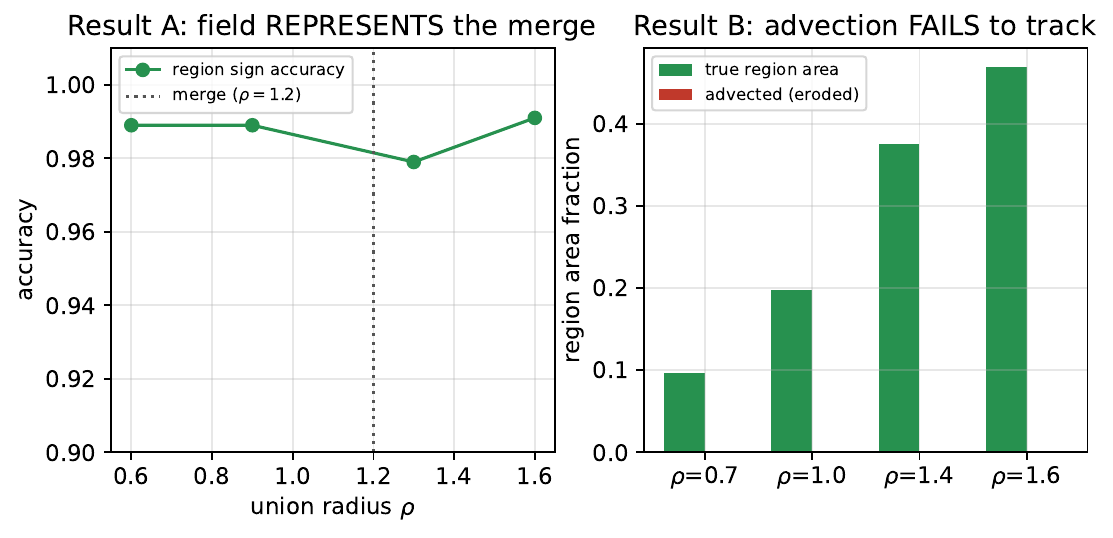}
\caption{Non-circular frontier. \textbf{Left (Result A):} the field correctly represents the
two-to-one topology change. \textbf{Right (Result B):} data-driven advection fails to track
it---the region erodes. The obstacle is the advection velocity on non-convex fronts, not the
field's representational capacity.}
\label{fig:limitation}
\end{figure}

The signature capability of a level-set representation---over an explicit boundary
parameterization---is handling frontiers that change \emph{topology}, e.g.\ two separate
consolidated regions that grow and merge. We test this with a two-seed benchmark where the
consolidated region is a union of two disks that merge when their radii exceed half the
center separation. Figure~\ref{fig:limitation} and Table~\ref{tab:topology} report two
findings.

\textbf{Result A (positive): representation.} The frontier field \emph{can represent} the
topology change: fitting $\phi_\psi$ to the union region yields correct connected-component
counts on both sides of the merge (two before, one after), region sign-accuracy
$0.98$--$0.99$, and a satisfied Eikonal constraint. Exact grid-based signed-distance
reinitialization is available and verified area-exact.

\textbf{Result B (negative): data-driven tracking.} The advection \emph{cannot yet} grow and
track this frontier through the merge: starting from two correct seeds, the advected region
erodes to near-zero area within one stage (Table~\ref{tab:topology}, right).

\begin{table}[H]
\centering
\begin{tabular}{lcc@{\hskip 2em}lcc}
\toprule
\multicolumn{3}{c}{Result A: representation} & \multicolumn{3}{c}{Result B: advection} \\
\cmidrule(r){1-3}\cmidrule(l){4-6}
$\rho$ & components & sign acc. & stage $\rho$ & area (true) & area (advected) \\
\midrule
$0.6$ & $2$ (\checkmark) & $0.99$ & $0.7$ & $0.097$ & $0.000$ \\
$0.9$ & $2$ (\checkmark) & $0.99$ & $1.0$ & $0.198$ & $0.000$ \\
$1.3$ & $1$ (\checkmark) & $0.98$ & $1.4$ & $0.375$ & $0.000$ \\
$1.6$ & $1$ (\checkmark) & $0.99$ & $1.6$ & $0.469$ & $0.000$ \\
\bottomrule
\end{tabular}
\caption{Topology-change study. \emph{Left:} the field represents the merge correctly
(Result A). \emph{Right:} data-driven advection erodes the region rather than tracking it
(Result B). The merge occurs at union radius $\rho=1.2$.}
\label{tab:topology}
\end{table}

\section{Scope and the Open Problem}
\label{sec:limitations}

The topology-change study in Section~\ref{sec:topology-results} delineates exactly what the
present method does and does not achieve, and we discuss its implications here.

The positive result (representation) and the negative result (data-driven tracking) together
localize the obstacle precisely. The failure in Result B is \emph{not} a representational
limitation---the field comfortably encodes both topologies---but a failure of the
\emph{velocity construction} of Eqs.~\eqref{eq:stefan}--\eqref{eq:advection}: the
closest-point normal $n=\grad\phi/\|\grad\phi\|$ is unstable near the \emph{medial axis}
between the two components, so the extended velocity becomes ill-defined in the concave
region and the front either erodes or spuriously bridges. This is the same construction that
succeeds on a single convex (radial) front, where reinitialization has a clean closed-form
target.

We therefore scope the present (v1) claims to frontiers admitting a clean reinitialization
target---radially symmetric or single convex components---and identify
\textbf{medial-axis-stable velocity extension} as the central problem for future work.
Promising directions: per-component velocities; PDE-based extension velocities solved on a
grid rather than via closest-point projection; or normals estimated by local smoothing
rather than by $\grad\phi/\|\grad\phi\|$. We regard the precise localization of this
obstacle---representation and reinitialization are solved; the velocity on non-convex fronts
is not---as a useful contribution in its own right.

\section{Conclusion}
Stefan-CL recasts the stability--plasticity dilemma as a moving-boundary problem: a learned
signed-distance frontier separates consolidated from plastic capacity, phase masks freeze
the interior, and a Stefan-condition velocity advects the frontier to consolidate new tasks,
with latent heat serving as a single calibrated stability--plasticity dial. On
analytically-grounded benchmarks the mechanism reduces forgetting $30\times$, recovers the
ground-truth growth law from data, beats regularization baselines, and matches replay
without stored data---all with tight error bars over $10$ seeds. We have been explicit about
scope: topology-changing frontiers are representable but not yet trainable, and the open
problem is a medial-axis-stable advection velocity. We hope the physics--ML mapping proves
generative beyond this first instantiation.

\section{Future Work}
\label{sec:future}

The present results open several concrete research directions, ordered roughly from the most
immediate (resolving the identified open problem) to the most speculative (extending the
physics analogy).

\subsection{Medial-axis-stable velocity extension}
The central open problem (Section~\ref{sec:limitations}) is that the closest-point velocity
extension destabilizes near the medial axis of non-convex fronts, preventing topology-changing
consolidation. Three avenues appear promising. First, \emph{per-component velocities}: detect
connected components of the consolidated region and extend a separate velocity field within
each, avoiding the ambiguous projection across the gap between components. Second, \emph{PDE-based
extension velocities}~\citep{adalsteinsson1999} solved on a background grid, which propagate the
interface velocity along normals by a transport equation rather than the algebraic closest-point
formula and are numerically stable in concave regions. Third, \emph{smoothed normal estimation},
replacing $n=\grad\phi/\|\grad\phi\|$ with a locally averaged or learned normal field that
degrades gracefully where $\grad\phi$ is ill-conditioned. Pairing any of these with the
fast-marching reinitialization~\citep{sethian1996} already validated here would directly test
whether the full multiphase mechanism becomes trainable.

\subsection{Higher-dimensional and learned frontiers}
Our benchmarks place the frontier in the 2D input space, where it can be visualized and checked
against analytic ground truth. The natural next step is to carry the frontier into a
\emph{learned latent space}: run a feature extractor, let $\phi$ live on its representation, and
advect there. This raises questions our toy setting cannot answer---how the Eikonal constraint
behaves in high dimension, and whether the frontier remains a meaningful object once the
representation itself drifts during training---and connects to representation-level continual
learning~\citep{javed2019}. Curvature-dependent interface speeds~\citep{osher1988} could additionally
regularize frontier geometry in high dimension.

\subsection{Task-free and online operation}
Stefan-CL currently consolidates at discrete task boundaries. Because the advection velocity is
driven purely by local data demand, the formulation is in principle \emph{task-agnostic}: the
frontier could advance continuously as data arrives, with no explicit task signal. Realizing
this would place Stefan-CL among task-free continual-learning methods~\citep{aljundi2019} and
remove a strong assumption shared with EWC and SI.

\subsection{Theory of the consolidation dynamics}
We observed empirically that the frontier radius is a near-deterministic function of the latent
heat under a fixed adaptation budget, and that it recovers the Frank-sphere law. A theoretical
account---deriving the protected-region growth rate from the data-demand velocity, and bounding
forgetting in terms of the protected fraction---would turn the latent-heat dial from a measured
relationship into a predictive one, and connect to statistical-physics analyses of learning
dynamics~\citep{shan2024,li2023}.

\subsection{Scaling and real benchmarks}
Finally, the mechanism must be validated beyond toy data on standard class- and
domain-incremental benchmarks~\citep{vandeven2022}, against stronger replay and parameter-isolation
baselines, and at the scale of modern architectures. The compactness of the frontier field
(a small auxiliary network) is encouraging for scaling, but the interaction between a moving
frontier and a high-capacity backbone is untested.

\bibliographystyle{plainnat}
\bibliography{refs}

\end{document}